\documentclass[10pt,twocolumn,letterpaper]{article}

\usepackage{cvpr}              
\usepackage{latexsym}
\usepackage{amssymb}
\usepackage{amsmath}
\usepackage{amsthm}
\usepackage{booktabs}
\usepackage{enumitem}
\usepackage{graphicx}
\usepackage{color}
\usepackage{arydshln}

\definecolor{cvprblue}{rgb}{0.21,0.49,0.74}
\usepackage[pagebackref,breaklinks,colorlinks,allcolors=cvprblue]{hyperref}

\def\paperID{*****} 
\def\confName{CVPR}
\def\confYear{2025}

\title{Towards Generating Realistic Underwater Images}

\author{Abdul-Kazeem Shamba\\
Marine and Maritime Intelligent Robotics\\
Norwegian University of Science and Technology\\
{\tt\small abdulkks@stud.ntnu.no}
}

\begin{document}
\maketitle
\begin{abstract}
This paper explores the use of contrastive learning and generative adversarial networks for generating realistic underwater images from synthetic images with uniform lighting. We investigate the performance of image translation models for generating realistic underwater images using the VAROS dataset. Two key evaluation metrics, Fréchet Inception Distance (FID) and Structural Similarity Index Measure (SSIM), provide insights into the trade-offs between perceptual quality and structural preservation. For paired image translation, pix2pix achieves the best FID scores due to its paired supervision and PatchGAN discriminator, while the autoencoder model attains the highest SSIM, suggesting better structural fidelity despite producing blurrier outputs. Among unpaired methods, CycleGAN achieves a competitive FID score by leveraging cycle-consistency loss, whereas CUT, which replaces cycle-consistency with contrastive learning, attains higher SSIM, indicating improved spatial similarity retention. Notably, incorporating depth information into CUT results in the lowest overall FID score, demonstrating that depth cues enhance realism. However, the slight decrease in SSIM suggests that depth-aware learning may introduce structural variations.
\end{abstract}    
\section{Introduction}
\label{sec:intro}

The creation of realistic synthetic underwater images provides numerous advantages for marine robotics practitioners and researchers \cite{lovaas2023semi, 7995024}. These images enable practitioners to test and optimize their specialized equipment in a controlled environment, ensuring accurate performance and functionality before deployment in real-world underwater settings. Additionally, practitioners in fields like underwater archaeology, marine biology, and oceanography can utilize these images for visualization and planning purposes, facilitating data interpretation and decision-making while ensuring safety and mitigating risks. Currently, the generation of realistic underwater images relies on complex physics-based models and extensive parameter adjustments, requiring domain knowledge of the underwater environment \cite{ alvarez2019generation}.

\begin{figure}[h]
\centering
\includegraphics[width=8.0cm]{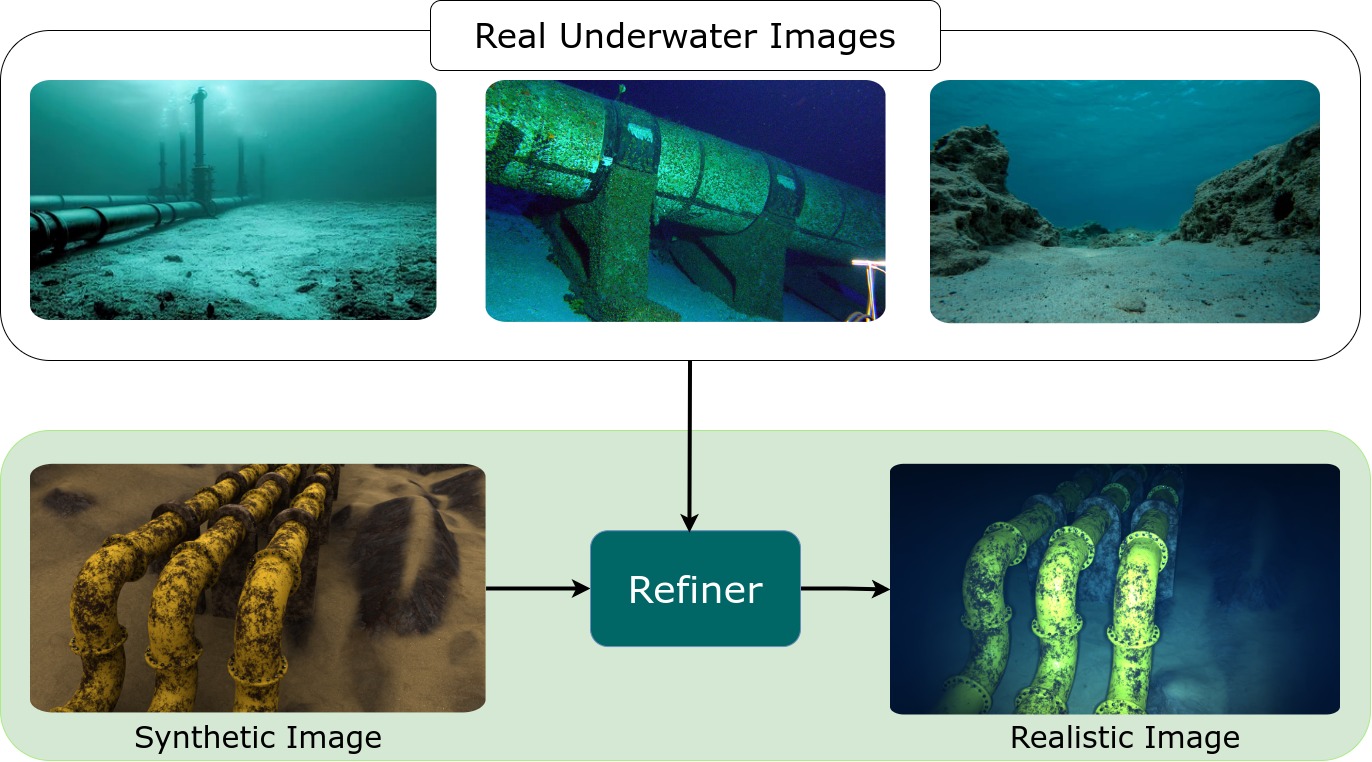}
\caption{Realistic underwater image generation. Given a synthetic image $X$ with uniform lighting and real underwater images $Y$, the unpaired image translation method learns to transform the synthetic input into a realistic underwater image.}
\label{fig:thesisdiag}
\end{figure}

This work explores state-of-the-art methods in contrastive learning and generative adversarial networks to generate realistic underwater images. This problem can be generalized as an image-to-image translation, converting an image of a scene from one representation $X$ to another $Y$. In this case, given a set of synthetic images with uniform lighting and another one of real underwater images, we aim to learn a model to map the appearance of the real underwater images to the content of the uniform lighting synthetic inputs (Figure \ref{fig:thesisdiag}). Previous work such as the use of ray tracing and physical models of the underwater environment \cite{blasinski2017underwater}, autoencoder-based image generation \cite{autoen, yu2023end},  neural style transfer non-photorealistic renderings \cite{gatys2016image} and paired image-to-image translation \cite{isola2017image} have proposed robust translation systems in the physics-based and supervised setting. However, paired training data can be expensive, particularly in controlled tank experiments, or impossible in some realistic underwater scenarios such as natural scenes and deep sea environments. Advancement in the domain of generative adversarial networks (GANs) \cite{goodfellow2014generative, karras2019stylebased, zhu2017unpaired}, contrastive representation learning \cite{sohn2016improved, chen2020simple}, and underwater image synthesis \cite{zwilgmeyer2021varos} form the foundation for the contrastive architecture in \cite{park2020contrastive}, which is an unpaired image-to-image translation. CycleGAN \cite{zhu2017unpaired} proposes the cycle-consistent loss to preserve the content across the two domains and tackle the issue of mode collapse in unpaired image translation using GANs. However, this approach requires two generators and discriminators, significantly slowing the training process. \cite{park2020contrastive} replaces the cycle-consistent loss in CycleGAN with a contrastive objective to enforce similarities of nearby patches and improve training time.

With the emergence of well-performing monocular depth estimation techniques and alternative methods for acquiring depth data, such as stereo vision, time of flight, and structured light, the process of incorporating depth into underwater scenes has become more feasible than ever before \cite{fu2018deep, kusupati2020normal, jiang2020time}. In addition to exploring the underwater image generation using paired and unpaired image translation techniques from state-of-the-art, this research aims to investigate the effectiveness of introducing depth information in the contrastive objective, specifically the CUT \cite{park2020contrastive}. We train the models on over 2000 synthetic images. We measure the performance on 190 uniform lighting images from the vision autonomous robots for ocean sustainability (VAROS) \cite{zwilgmeyer2021varos} synthetic underwater dataset using Fréchet inception distance (FID) \cite{heusel2017gans} and (SSIM) \cite{wang2004image} evaluation metrics. Our findings suggest that contrastive representation learning with depth information and GAN generates realistic underwater images that can rival the physics-based methods while using significantly lesser training time and memory than other unpaired image translation methods such as CycleGAN \cite{zhu2017unpaired}. This research makes two main contributions. Specifically:


\begin{itemize}
    \item Conduct experiments to investigate state-of-the-art paired and unpaired image translation using contrastive and GAN objectives in underwater image generation.
    
    \item Introducing depth information in the contrastive learning objective of CUT \cite{park2020contrastive} in underwater image generation that requires reflectance and lighting understanding produces high-fidelity images (Figure \ref{fig:sixtest}).
       
\end{itemize}


\section{Related Works}
\label{sec:formatting}

\textbf{Physics-based image synthesis.}
Underwater image formation is a complex process influenced by reflectance, backscatter, light propagation, and refraction. Understanding the underwater image formation model (IFM) is critical to creating realistic underwater renderings. The key difference with in-air image formation is the strong wavelength-dependent attenuation and scattering. In waters beyond a few meters deep, nearby artificial lighting is needed, which can make backscatter effects worse. Prior works have made progress in creating realistic-looking underwater renderings, as in deep sea underwater simulators \cite{sedlazeck2011simulating}. \cite{alvarez2019generation} proposed a framework that effectively combines post-processing by an unmanned underwater vehicles simulator \cite{manhaes2016uuv} with the linearized formula suggested in \cite{jaffe1990computer} and style transfer technique to create realistic underwater images. More recent work by \cite{song2021deep} presents a physical model-based solution that combines in-air texture and depth information to generate deep sea underwater image sequences. The authors acknowledge the effect of artificial illumination in the deep sea IFM to account for both attenuation and backscattering effects. Other works \cite{zwilgmeyer2021varos, blasinski2017underwater} aim to create a highly realistic scene rendering using the ray tracing method while integrating direct light and indirect volumetric scattering. All the methods mentioned above require domain knowledge of the underwater properties and parameters present in the IFM.
\\

\noindent\textbf{Autoencoder image generation.} The limitations of the previous physics-based method have spurred researchers to develop end-to-end networks that automatically learn to generate underwater images. Autoencoder-based image reconstruction forms the backbone for some of the early works in this area. \cite{autoen, fabbri2018enhancing} use the U-Net denoising autoencoder for underwater color restoration. \cite{yu2023end} also proposed an end-to-end underwater image enhancement framework that combines fractional integral-based retinex using an unsupervised autoencoder network that integrates an advanced attention mechanism. Recently, \cite{kim2021pixel} introduced a dehazing approach using the Wasserstein autoencoder. Compared to a conventional autoencoder with low dimensional latent vectors, their approach learns a 2-dimensional latent tensor and matches the haze inputs with the dehazed output pixel-wise. Autoencoder sometimes produces ambiguous reconstructions that occupy different distributions from the input, hence the reason for the development of variational autoencoder \cite{kingma2013auto} to impose structure on the latent space. \cite{xu2019adversarially} proposed a novel unregularized adversarially approximated autoencoder to approximate the implicit probability distribution. The methods described above are all used for dehazing or enhancing underwater images, which differs from the task in this work. 
\\


\noindent\textbf{Neural style transfer.} Style transfer has its origin in non-photorealistic renderings and is closely related to texture synthesis and transfer. Neural style transfer is a technique that effectively combines the content of one image with the style of another to create a new image. The fundamental idea behind style transfer is that an image style and content are separable entities \cite{gatys2016image}. In an underwater setting, the content of the image can be the features of the image as it would appear in-air, such as the terrain, corals, and reefs, while the style would factor in the color, backscatter, and other optical parameters that are manipulated using the IFM. However, applying gradient-based optimization on the image space during inference is limited and inefficient, especially in real-time applications such as applying new styles to video. We want to learn a model that can instead efficiently apply an underwater realistic style to a sequence of images in real time. Neural style transfer has seen good applications in various domains, such as artistic renderings and underwater. \cite{lee2019deep} applied style transfer on underwater sonar images given a depth image obtained from a simulator while optimizing the losses in \cite{johnson2016perceptual}. \cite{zhou2021matching} designed a method that uses the style transfer technique by converting sonar images into optical styles to enhance the matching performance of underwater sonar images. Other works, such as \cite{karras2019stylebased}, build upon the idea of neural style transfer to develop a GAN-based model. \\

\noindent\textbf{Image-to-image translation.} Going beyond neural style transfer, GANs \cite{goodfellow2014generative} has seen some remarkable applications in image-to-image translation. Advancement in GANs and the work by \cite{gatys2016image} has inspired several works, such as StyleGAN \cite{karras2019stylebased}, StarGAN \cite{choi2018stargan}, and CycleGAN \cite{zhu2017unpaired}. In the underwater environment, GAN is commonly used to create a realistic underwater dataset for training end-to-end image enhancement model \cite{li2018synthesis, li2017watergan, shrivastava2017learning}. The pix2pix \cite{isola2017image} is a powerful paired model that uses conditional GAN to learn a mapping function from the input image to a target image. To achieve unpaired image-to-image translation, CycleGAN learns a mapping function to translate an image from a source
domain to a target domain in the absence of paired examples. CycleGAN uses two generators and two discriminators to translate images in two domains and proposes the cycle-consistent loss to tackle the issue of mode collapse in unpaired image translation. Finally, a more recent work introduces CUT \cite{park2020contrastive}, which replaces the cycle-consistent loss in CycleGAN with a contrastive objective to enforce similarities of nearby patches and reduce training time.

\section{Generating Underwater Images}

The task of generating realistic underwater images from synthetic uniform lighting renderings is seen as translating an image from one domain to another. In this work, the two domains are uniform lighting synthetic images and realistic underwater images. The goal is to learn a mapping function that can translate images from the uniform lighting domain $X \subset \mathbb{R}^{H \times W \times 4}$ to the style of an image from the underwater domain $Y \subset \mathbb{R}^{H \times W \times 3}$. We are provided with a dataset of unpaired instances $X = \{x \in X\}$, $Y = \{y \in Y\}$.

\begin{figure}[h]
\centering
\includegraphics[width=7.8cm]{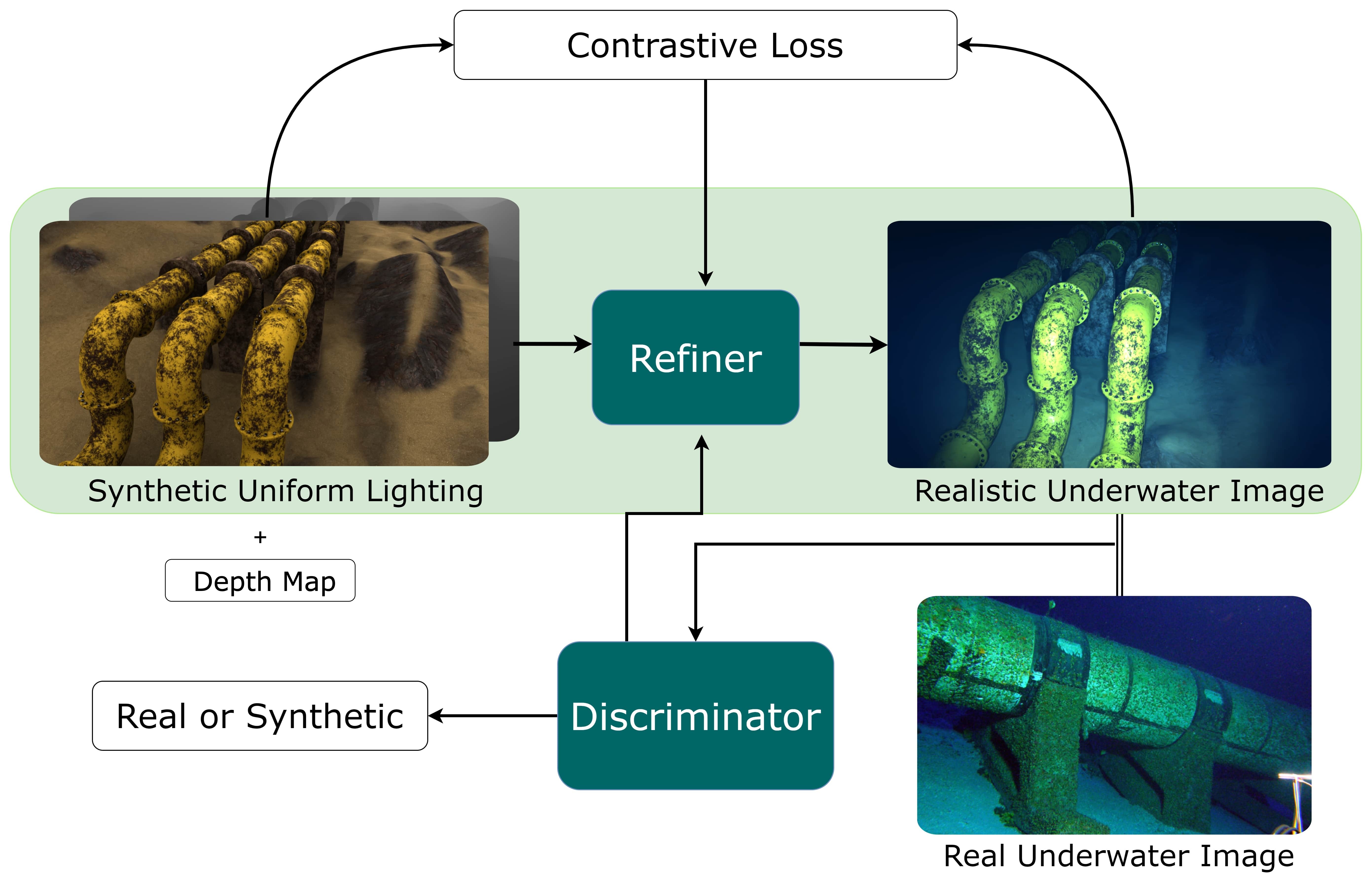}
\caption{A realistic underwater image generation learning procedure using contrastive objective and depth (CUT + \textit{depth}). The refiner is the model generator conditioned on real underwater images. The contrastive loss preserves the content of the synthetic uniform lighting input \cite{park2020contrastive}.}
\label{fig:codealg}
\end{figure}

Image-to-image translation has been chiefly solved using two techniques -- paired image-to-image translation and unpaired image-to-image translation. We explore both methods in this work. First, we build a somewhat naive image translation model using an autoencoder and compare the results with those obtained using the state-of-the-art pix2pix \cite{isola2017image}. Second, we adapted the CUT \cite{park2020contrastive} framework and adjusted the architecture to account for an extra dimension of depth information to enforce underwater reflectance and lighting understanding in the model (see Figure \ref{fig:codealg}). The core of this research is to investigate both generative techniques for creating realistic underwater images from synthetic uniform lighting images. The architecture is made up of a refiner, $R_\theta(.)$, which generates realistic underwater images, and a discriminator, $D_\phi(.)$, that attempts to distinguish the generated images from real underwater images. The contrastive loss from CUT \cite{park2020contrastive} ensures that the corresponding patches from the input and output images occupy similar embedding space. This is necessary to preserve the content, which needs to remain consistent across domains.

\subsection{Adversarial and Contrastive Objectives}

We train the refiner network $R_\theta(.)$ to generate realistic underwater images that make it difficult for the discriminator $D_\phi(.)$ to distinguish from the real ones. CUT \cite{park2020contrastive} split the refiner network $R_\theta(.)$ into two components, an encoder $R_{\theta,\text{enc}}$ followed by a decoder $R_{\theta,\text{dec}}$, where $z = R_{\theta,\text{enc}}(x)$ and generated output $\hat{y} = R_{\theta,\text{dec}}(z)$. Separating the refiner $R_\theta(.)$ into two components allows the authors to reuse the trained encoder $R_{\theta,\text{enc}}$ to map the input $x$ and output $\hat{y}$ into an embedding space $z$ for use in the patchwise contrastive learning. The refiner network $R_\theta(.)$, which maps image from $x$ to output $\hat{y}$ while preserving the content of domain $X$, is trained simultaneously with the discriminator $D_\phi(.)$, conditioned on underwater images $Y$. Where $R_\theta(.)$ minimizes the objective function against an adversary $D_\phi(.)$ that aims to maximize the objective function.

\begin{equation}\label{eq:vcg}
    \begin{aligned}
        \mathcal{L}_{\mathrm{GAN}}(R, D, X, Y) & =\mathbb{E}_{y \sim p_{\text {data }}(y)}\left[\log D_\phi(y)\right] \\
        & +\mathbb{E}_{x \sim p_{\text {data }}(x)}\left[\log \left(1-D_\phi(R_\theta(x))\right]\right .
    \end{aligned}
\end{equation}

\begin{figure}[h]
\centering
\includegraphics[width=7.8cm]{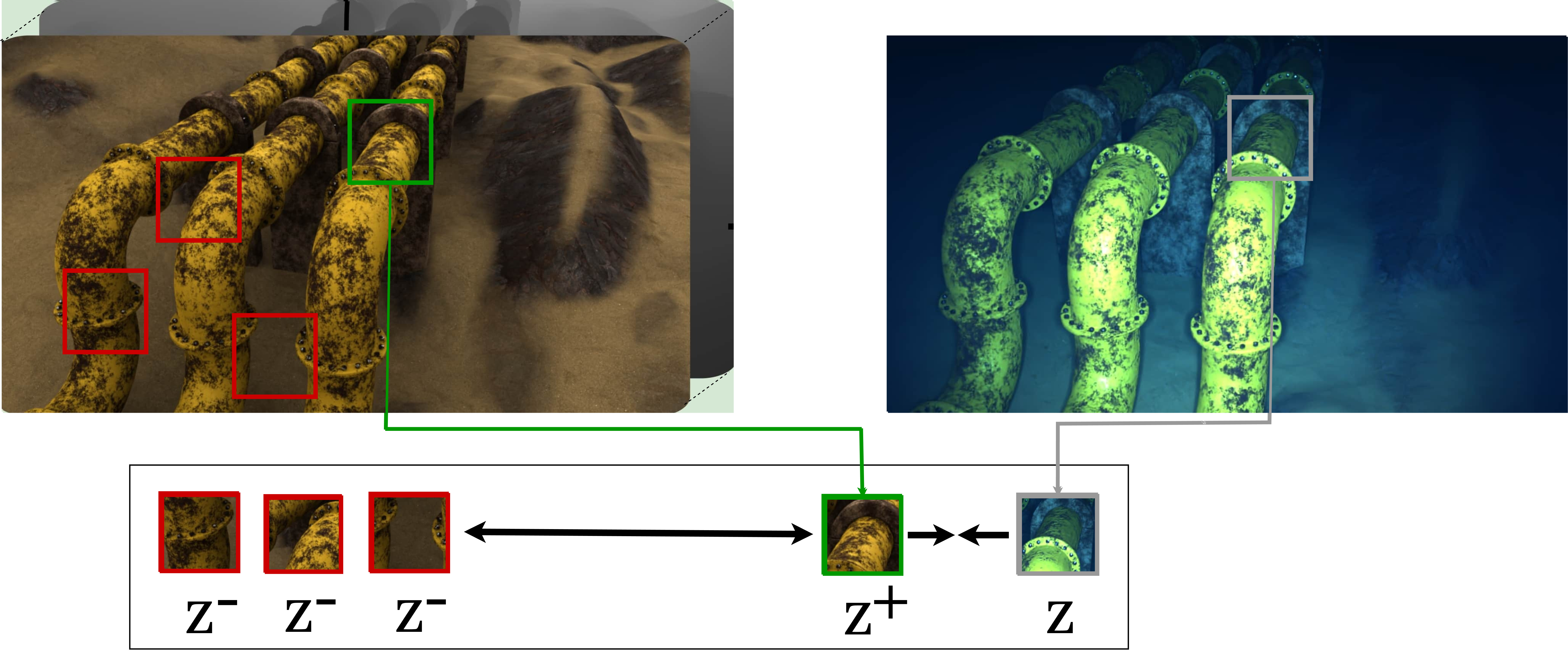}
\caption{Patchwise contrastive learning for generating realistic underwater images. Generated output patches $z$ and the corresponding input patch $z^+$ should occupy similar embedding spaces, and patches from random locations $z^-$ are pushed apart \cite{park2020contrastive}.}
\label{fig:patch}
\end{figure}

Implementing the vanilla GAN objective above leads to mode collapse.  In mode collapse, the generator network in the GAN fails to capture the full diversity of the training data and instead only learns to generate a limited subset of samples. CycleGAN \cite{zhu2017unpaired} introduced two generators and two discriminators to tackle the problem of mode collapse and to reduce the space of possible mapping functions. $G: X \rightarrow Y $ for mapping $X$ to $Y$ and $F: Y \rightarrow X$ for mapping $Y$ to $X$. We extend the method from \cite{park2020contrastive}, which uses patchwise contrastive loss. This method only requires a single generator and discriminator, which removes complexities from the training procedure and reduces training time. \cite{park2020contrastive} revisit the InfoNCE loss with a scaled temperature parameter $\tau$  to map nearby patches to similar embedding spaces \cite{sohn2016improved, chen2020simple}. The uniform lighting images and the generated images are divided into patches; then, a contrastive loss is applied to these patches. The anchor is a patch from the generated image, and the corresponding patch in the uniform lightning input is the positive patch. Furthermore, the negative patches are selected as samples from the same uniform lighting input (Figure \ref{fig:patch}). The final loss set up is an N-pair loss objective\cite{sohn2016improved}. The anchor, positive, and N negatives are each mapped to K-dimensional vectors $z$, $z^+$ $\in \mathbb{R}^K$, and $z^{-}$ $\in \mathbb{R}^{N \times K}$, respectively. $z^{-}_n$ $\in \mathbb{R}^K$ represents the vector for the n-th negative.



\begin{equation}
\ell(\boldsymbol{z}, \boldsymbol{z^+}, \boldsymbol{z^-}) = 
- \log \frac{\exp(z \cdot z^+/\tau)}{\exp(z \cdot z^+/\tau) + \sum_{n=1}^{N} \exp(z \cdot z^-_n /\tau)}
\label{eq:patch}
\end{equation}



CUT reuses the encoder part of the refiner network to map input $x$ and output $\hat{y}$ into a feature space $z$. The encoder, denoted $R_{\theta,\text{enc}}$, consists of feature stacks where each layer and spatial location within the stack represents a patch of the input image.

The PatchNCE loss proposed by \cite{park2020contrastive} involves selecting $L$ layers of interest. Each feature map from these layers is passed through a two-layer MLP network $H_l$ to produce a stack of features:
\[
\{z_l\}_{l=1}^{L} = \{H_{l}(R_{\theta,\text{enc}}^{l}(x))\}_{l=1}^{L},
\]
where $R_{\theta,\text{enc}}^{l}$ denotes the output of the $l$-th layer. Layers are indexed as $l \in \{1, 2, \ldots, L\}$, and spatial locations within each layer are indexed by $s \in \{1, \ldots, S_{l}\}$. The corresponding feature is $z^{s}_{l} \in \mathbb{R}^{C_{l}}$, while the other features in the same layer are denoted $z^{S \setminus s}_{l} \in \mathbb{R}^{(S_{l}-1) \times C_{l}}$, where $C_{l}$ is the number of channels at layer $l$.

Similarly, the output image $\hat{y}$ is encoded as:
\[
\{\hat{z}_l\}_{l=1}^{L} = \{H_{l}(R_{\theta,\text{enc}}^{l}(\hat{y}))\}_{l=1}^{L}.
\]

The PatchNCE loss is defined as:

\begin{equation}
    \mathcal{L}_{\text{PatchNCE}}(R, H, X) = \mathbb{E}_{\boldsymbol{x} \sim X} \sum_{l=1}^{L} \sum_{s=1}^{S_l} \ell\left(\hat{\boldsymbol{z}}_l^s, \boldsymbol{z}_l^s, \boldsymbol{z}_l^{S \setminus s}\right)
    \label{eq:patch}
\end{equation}


\textbf{Combined objective function.} The final objective in CUT \cite{park2020contrastive} includes two terms: (1) an adversarial loss ensuring that generated images resemble those from the target (underwater) domain, and (2) the PatchNCE loss, which ensures that corresponding patches from input and output images occupy similar embedding spaces, preserving content across domains.


\begin{equation}
    \begin{aligned}
        \mathcal{L} & = \mathcal{L}_{\mathrm{GAN}}(R, D, X, Y) + \mathcal{L}_{\text {PatchNCE }}(R, H, X) \\
        & + \mathcal{L}_{\text {PatchNCE }}(R, H, Y).
    \end{aligned}
\end{equation}


\subsection{Model architecture}

We use a generative adversarial network with its generator and discriminator architecture for the unpaired image translation task similar to \cite{park2020contrastive}. We adjusted the generator network to process 4-channel input, which includes depth information. In this work, we refer to this modified generator as the "refiner.". The core components of the refiner and discriminator architecture are discussed in the following subsections.
\\

\noindent\textbf{Refiner.} The input to the refiner network is a 4-channel image. The network contains a series of normal fractional strided convolutions and several residual blocks. We use 9 blocks of residual layers in this experiment. All non-residual layers, except the output layer, follow batch normalization and the ReLU activation function. The output of the refiner is a realistic underwater image. At the end of the training procedure, the refiner network is retained and used for image generation during inference.
\\

\noindent\textbf{Discriminator.} We use the PatchGAN developed by pix2pix \cite{isola2017image} for the discriminator architecture. The idea behind PatchGAN is to restrict the focus to smaller patches that capture high-frequency structures in an image. PatchGAN only penalizes structure at the scale of small patches. A binary classification of real or fake is performed on all patches of size $n \times n$ patches, and all outputs are averaged to produce the overall discriminator output.

\subsection{Training data}

We use the synthetic VAROS dataset \cite{zwilgmeyer2021varos} in training the adversarial and autoencoder network.
\\

\noindent\textbf{Paired image translation.} We extracted 751 uniform lighting and their paired underwater equivalent from the VAROS dataset. We optimize the autoencoder model with the mean squared error (MSE) loss between the generated images and the expected ground truth, in this case, the equivalent underwater images from the VAROS dataset. The model takes an image as input and converts it to a tensor at a spatial resolution of 64. At each iteration of the training procedure, the model processes a batch of uniform lighting images and the corresponding underwater ground truth images for MSE loss computation. We use the same arrangement of data for training the pix2pix model. 
\\

\begin{figure}[h]
\centering
\includegraphics[width=7.8cm]{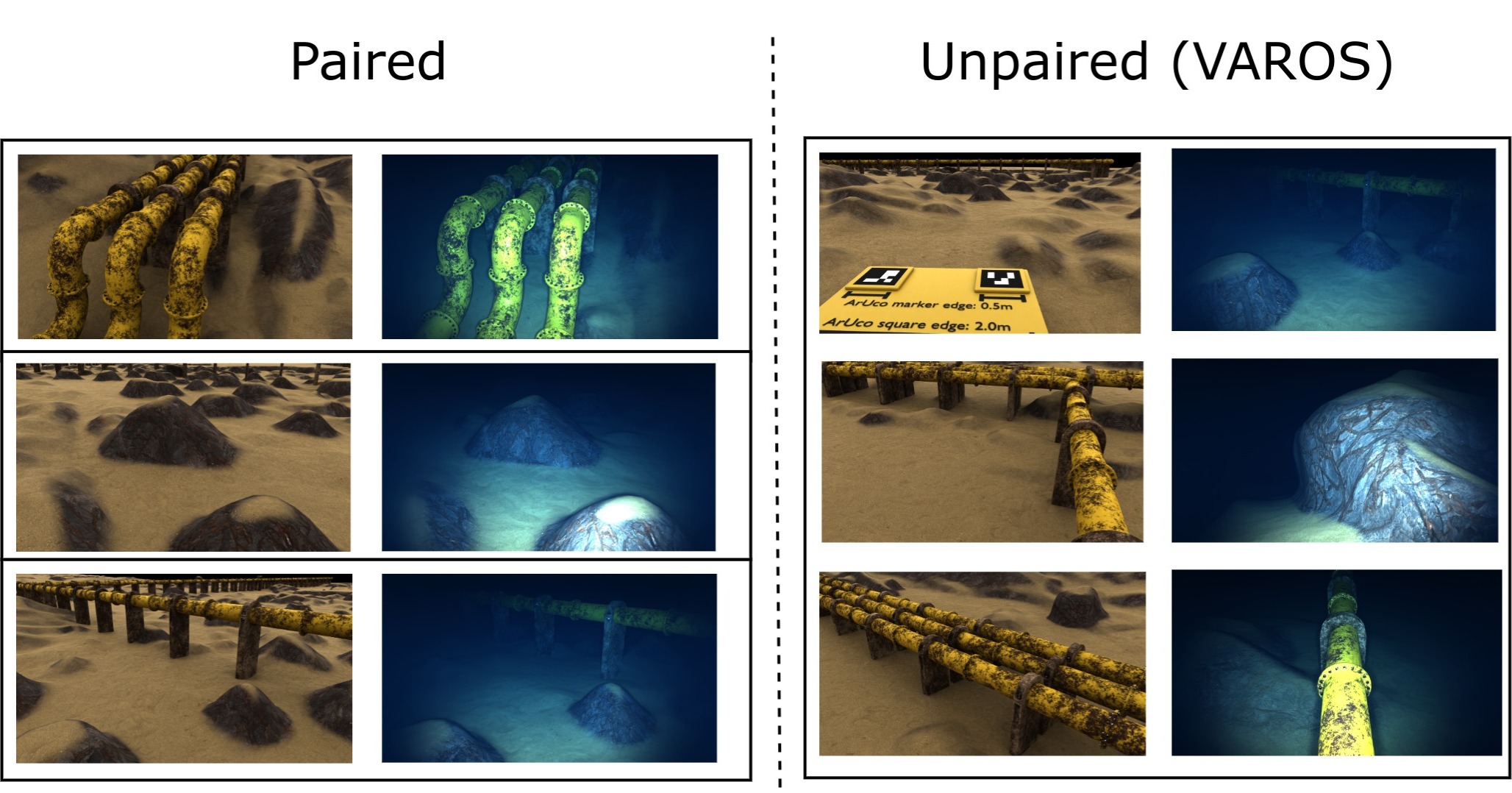}
\caption[Paired and unpaired training data setup]{Paired training data (left) consists of training examples $\{x_i, y_i\}$, where the correspondence between $x_i$ and $y_i$ exists from folders B and A in the VAROS synthetic underwater dataset respectively. We consider unpaired training data (right), consisting of a source set $\{x_i\}$ ($x_i \in X$) and a target set $\{y_j\}$ ($y_j \in Y$) sampled from synthetic underwater images, with no information provided as to which $x_i$ matches which $y_j$. Unpaired training example by using the synthetic input $x_i$ for $i=1$ to $K$ from VAROS and a target set $y_j$ for $j=K-N$ to ensure no correspondence.}
\label{fig:datasetup}
\end{figure}

\noindent\textbf{Unpaired image translation.} The overarching objective of unpaired image-to-image translation involves transforming synthetic images with uniform lighting into realistic underwater scenes. This task involves using a GAN refiner model conditioned on real underwater images. Given that real underwater images lack direct correspondence in content and structure with the input synthetic images, we classify this task as an unpaired image-to-image translation. We created the unpaired dataset as seen in Figure \ref{fig:datasetup}. We selected 1090 uniform lighting images corresponding to sequences 1011 to 2100 from the VAROS synthetic dataset. For the underwater images, instead of selecting a corresponding sequence as in the case of paired image translation, we rather select unpaired 1011 images corresponding to sequence 0000 to 1011.

\subsection{Model evaluation}

After training, we extracted the refiner network and evaluated it on the test images. The test images are selected from 190 uniform lighting images from the VAROS dataset \cite{zwilgmeyer2021varos}, to investigate the robustness of the refiner network; specifically, we check the realism of the generated images. Evaluating the realism of the output-generated images from autoencoder and GAN is an open and challenging problem. Aside from assessing the visual quality of the generated data by subjective human evaluation, we also compute the FID and SSIM between the generated output and the expected underwater ground truth. We follow standard practice in evaluating GAN images by reporting the FID scores, which compute the divergence between the estimated distribution of real and generated images. Ideally, realistic images should have a similar distribution to real underwater images regardless of the feature space. We report the FID and SSIM scores for the paired and unpaired image-to-image translation in Table \ref{tab:eval}. Additionally, we show the FID and SSIM score for the six randomly selected images in Figure \ref{fig:sixtest}.
\section{Experiments}

In this section, we perform comprehensive experiments and explore different design choices, such as model selection, dataset preprocessing, and training parameters. We divide the experiments into two categories - paired and unpaired image translation. We build an end-to-end autoencoder model for the first category and compare the results with the state-of-the-art in paired image translation pix2pix \cite{isola2017image}. Similarly, for the unpaired image translation, we explore the underwater generative capabilities of established baselines \cite{li2017watergan, shrivastava2017learning, zhu2017unpaired}, and extend \cite{park2020contrastive} to account for depth information.

\subsection{Training paired image translation}

\begin{figure}[h]
\centering
\includegraphics[width=7.8cm]{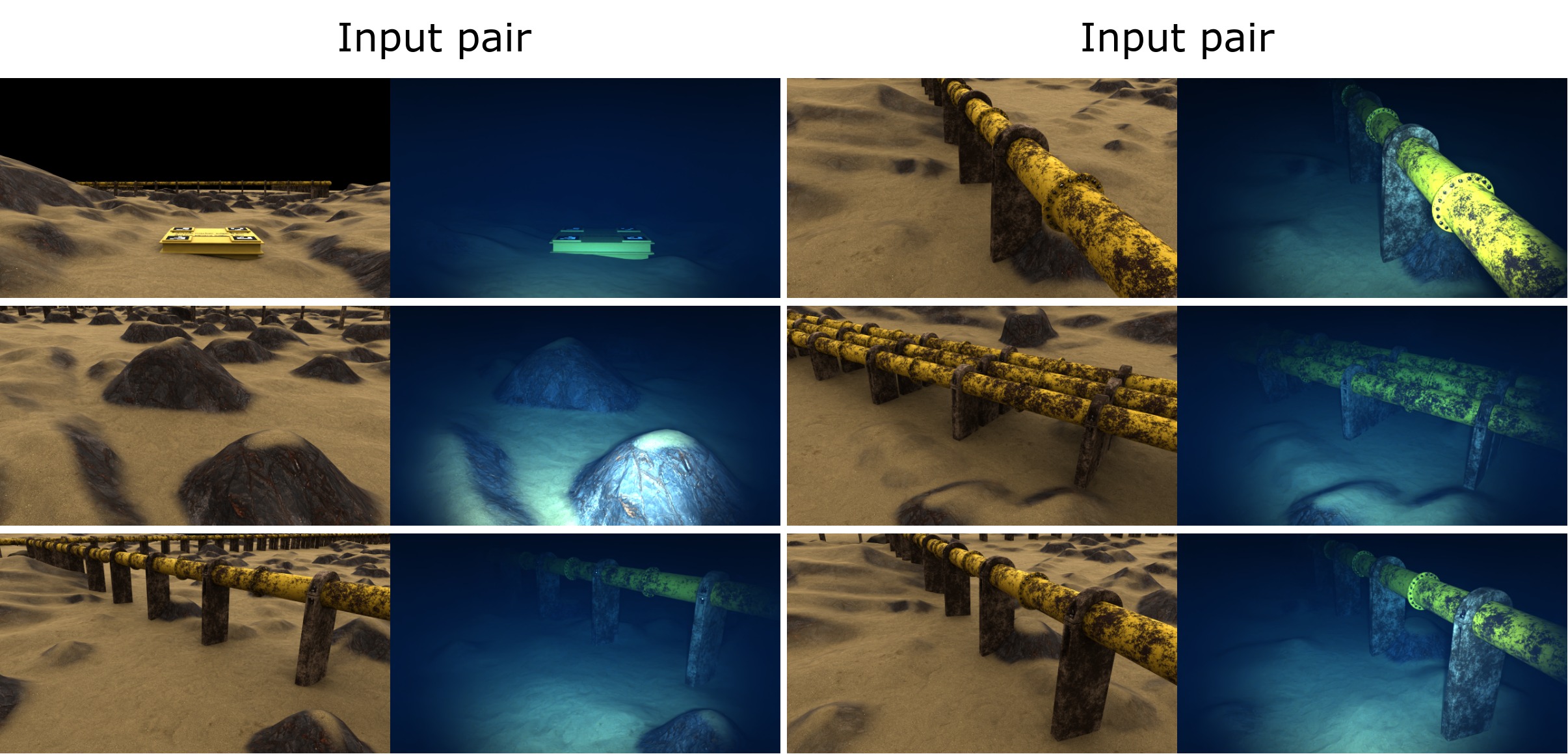}
  \caption[Paired training data setup for the autoencoder model]{Paired training data (left) consists of training examples $\{x_i, y_i\}$, where the correspondence between $x_i$ and $y_i$ exists from folders B and A in the VAROS synthetic underwater dataset respectively, for training the autoencoder and pix2pix model.}
  \label{fig:pixdata}
\end{figure}

\textbf{Data setup.} We first set up the data to train the autoencoder and pix2pix model. We select 751 consecutive scenes as training examples from the uniform lighting and the corresponding underwater images to train the autoencoder and pix2pix model. We evaluate all models, irrespective of their category, on randomly selected test images containing diverse scenes. The training set is selected from the VAROS environment and placed in two folders. The uniform lighting and underwater sequence are sufficient for training the autoencoder model with the custom dataloader. However, for the pix2pix, further processing is needed. The pix2pix needs the input pair to be stacked side by side in a single folder (Figure \ref{fig:pixdata}). To achieve this, first, we need to ensure corresponding image sequences have the same name. Next, we concatenate this sequence horizontally to form a new folder with 751 images.

\noindent\textbf{Network.} We explored two main model architectures for the paired image translation -- the autoencoder network and the pix2pix network. For this work, we use the encoder-decoder architecture and compare this with the baseline pix2pix \cite{isola2017image}. The autoencoder network comprises of convolutional, transpose convolutional networks, and the feature extraction head consists of the  ResNet34 architecture \cite{he2015deep}. The pix2pix model comprises two networks -- the image generation backbone uses U-Net256 \cite{ronneberger2015unet}, and the PatchGAN \cite{isola2017image} discriminator architecture enforces underwater realism in the model output.
\\
\noindent\textbf{Training details.} We train the autoencoder model with the following hyperparameters: learning rate (lr) $1e^{-3}$; we apply the Adam optimizer with a weight decay of $5e^{-5}$. In addition, we use the MSE loss as the optimization criterion. We trained this network for 500 epochs on an RTX 3090 GPU with 128GB RAM. We log the training metric each time, track the training procedure, and generate an image on the weights and biases (WandB) platform. On a GPU, the autoencoder model trains for an average of 40 seconds per epoch and 12 minutes for each epoch on a CPU. The input to the model is a 256 x 256 RGB image.

\begin{figure}[h]
\centering
\includegraphics[width=8.0cm]{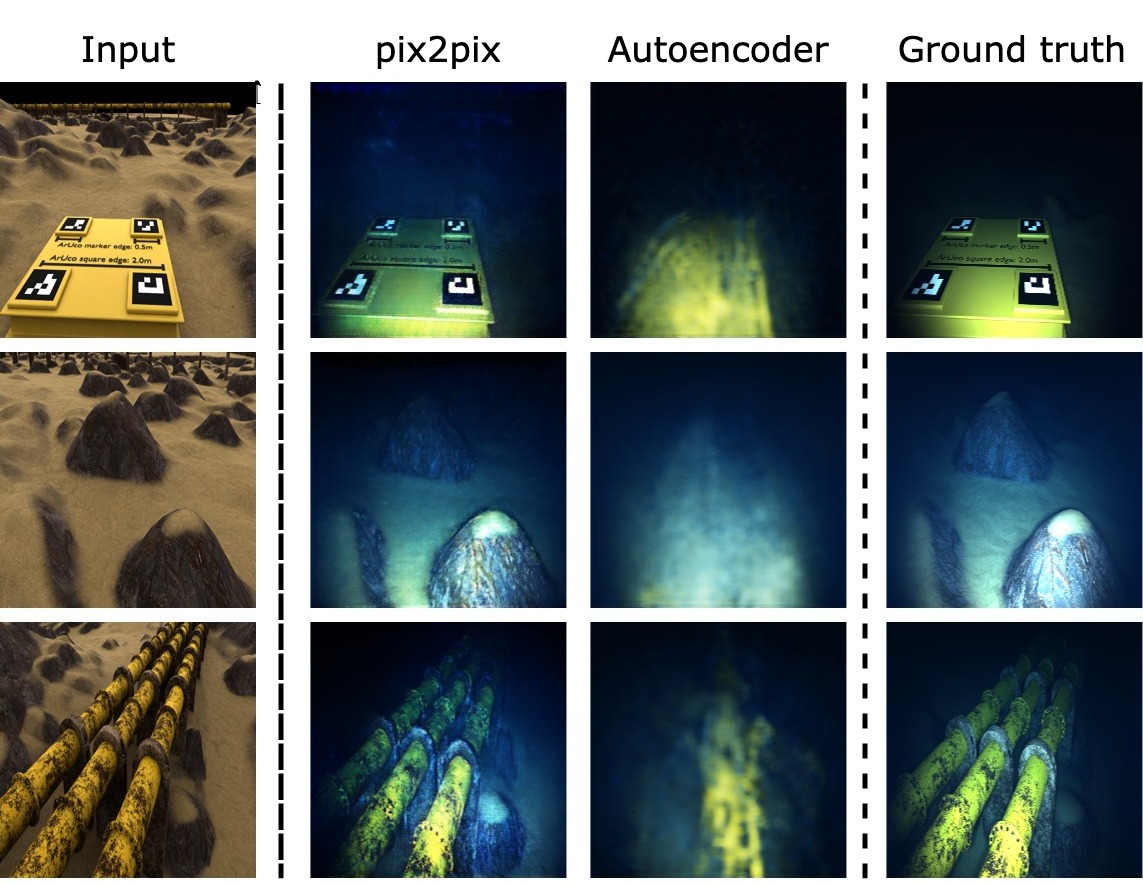}
  \caption{Results generated using pix2pix and autoencoder paired image translation models. The images in the first column are three examples from the VAROS synthetic datasets.}
  \label{fig:basecom}
\end{figure}

\subsection{Training unpaired image translation}

\textbf{Data setup.} To achieve unpaired image-to-image translation, we structure the VAROS dataset to encapsulate the two domains we aim to capture in this work. We created an unaligned pair with the VAROS dataset. This unpaired sequence of the VAROS dataset underpins the ability of the models to learn a mapping F that effectively translates images from one domain to another without paired training examples. Unlike the paired image translation, we added the depth images in the data processing. We intentionally extract unequal numbers of images for each domain. For example, we selected 1011 and 1090 images for synthetic underwater and uniform lighting, respectively. We did this to highlight that this is an unpaired image translation dataset with no explicit need for the number of training images in each category to be the same.
\\

\noindent\textbf{Network.} We adapted the architecture provided in \cite{park2020contrastive} for the unpaired image translation and extended the network to account for the extra depth channel in the VAROS dataset. The goal is to use the depth information to force CUT \cite{park2020contrastive} model to learn underwater light attenuation. 
\\

\noindent\textbf{Training details.} The hyperparameters of the CUT \cite{park2020contrastive} network are the same as those used in CycleGAN \cite{zhu2017unpaired} except that the loss function is replaced with contrastive loss. We train the CUT + \textit{depth} using 256 x 256 4-channel RGBD images with 9 residual blocks, PatchGAN discriminator, least square GAN loss, Adam optimizers, lr $2e^{-3}$, and a batch size of 8. We train all networks for 200 epochs.


\section{Underwater Image Translation}

We begin by investigating the realism of all state-of-the-art methods for unpaired image translation. We then performed an ablation study to understand the relevance of depth maps in generating realistic underwater images. Lastly, we explore the explainability of the autoencoder model by performing additional experiments to extract embeddings from different layers. For a fair comparison, we train all baselines for similar epochs and use similar parameters where possible. We divide the baselines into two categories -- trained on paired data and trained on unpaired data.

\subsection{Methods for Image Translation}

\noindent\textbf{pix2pix \cite{isola2017image}} This model is trained on paired data and is compared to the autoencoder model. Also, to check the robustness of the unpaired image-to-image translation models, we compared the results with that of pix2pix.

\noindent\textbf{CycleGAN \cite{zhu2017unpaired}} This method uses two generators and discriminators and a cycle-consistent loss to translate an image from a source domain $X$ to a target domain $Y$ using unpaired training data.

\noindent\textbf{CUT \cite{park2020contrastive}} This work improves the CycleGAN by replacing the cycle-consistent loss with a contrastive loss.

Figure \ref{fig:basecom} shows three examples obtained using the autoencoder-based image generation and pix2pix \cite{isola2017image}. The pix2pix produces more fine-grained images than the autoencoder-based image generation. This can be attributed to the fact that pix2pix uses the PatchGAN discriminator to encourage high-frequency crispness in an image. In contrast, the autoencoder uses traditional metrics like the MSE loss, which focuses on low-level pixel similarity while ignoring overall structural information, thus leading to blurry results \cite{larsen2016autoencoding}. We provide the quantitative comparison with the ground truth by computing FID and SSIM scores in Table \ref{tab:eval}.

\begin{table}[h]
    \centering
        \caption[Comparison with baselines using FID and SSIM metrics]{\textbf{Comparison with baselines}. We compare baseline methods on the VAROS underwater datasets with the FID and SSIM evaluation metrics. We show FID, a measure of image quality \cite{heusel2017gans} (lower is better), and the SSIM (closer to 1 is desired).}
        \resizebox{0.48\textwidth}{!}{ 

        \tiny 
        \begin{tabular}{lcccc}
        \toprule
        &Samples (6) && Samples (190) \\
        \cmidrule(r){2-3}\cmidrule(lr){4-5}
        Method \hspace{1cm} & SSIM$\uparrow$ & FID$\downarrow$ & SSIM$\uparrow$ & FID$\downarrow$\\
        
        \midrule
        Pix2Pix & 0.66 &  \textbf{206} & 0.68 & 128\\
        Autoencoder & \textbf{0.70} & 314 & 0.67 & 268\\
        \hdashline
        CycleGAN & 0.61 & 241 & 0.67 & 108\\
        CUT & 0.62 & 299 & \textbf{0.69} & 132\\
        CUT + \textit{depth} & 0.57 & 278 & 0.66 & \textbf{106}\\
        \bottomrule
        \end{tabular}
     }
    \label{tab:eval}
\end{table}

\begin{figure}[h]
\centering
\includegraphics[width=7.5cm]{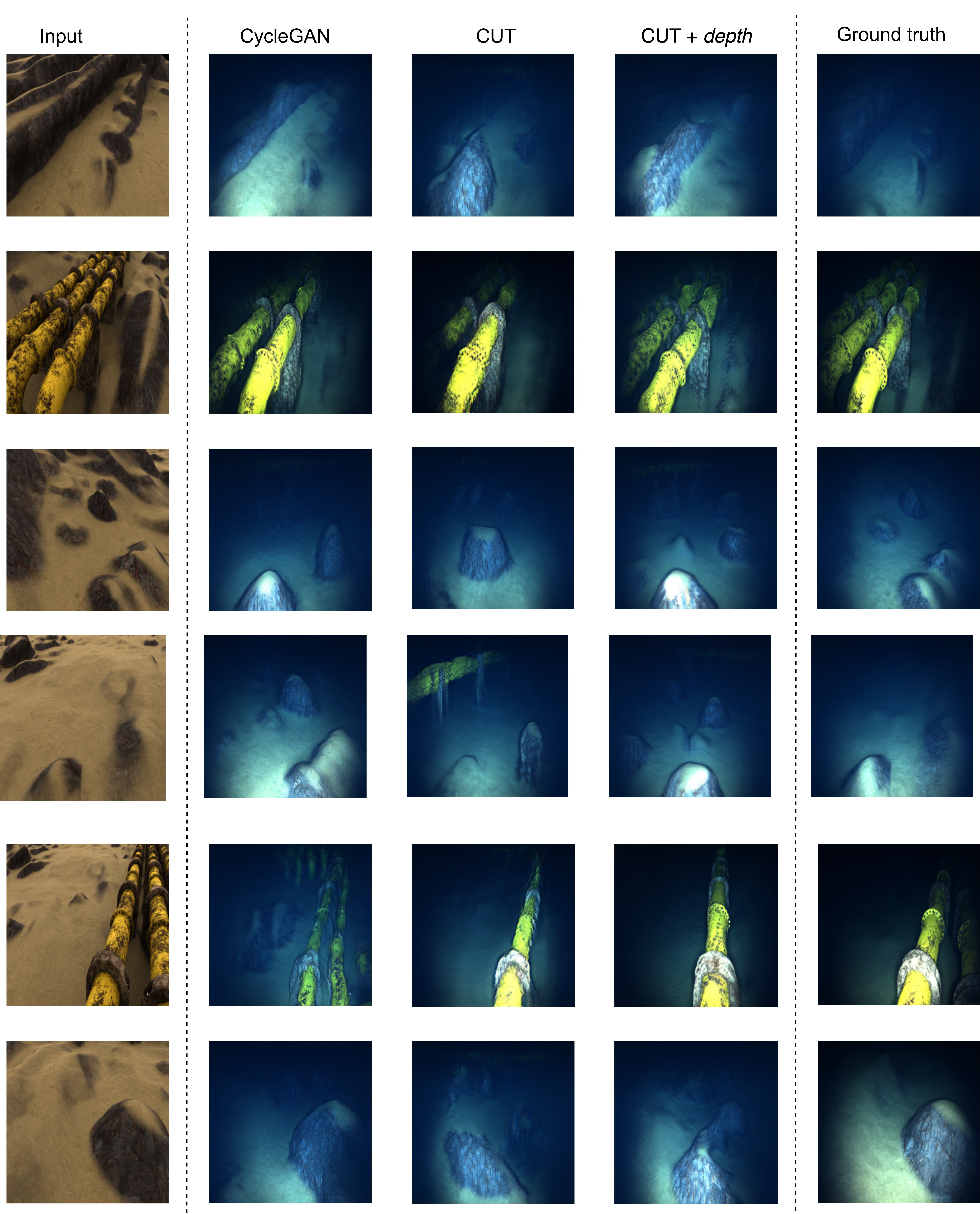}
  \caption{Results from different baselines from six randomly selected images in the test set.}
  \label{fig:sixtest}
\end{figure}

We provide the results on inference from the six random samples in Figure \ref{fig:sixtest} and the entire 190 test images. The inference on 6 samples underscores the FID as a desired metric as opposed to SSIM, which ranks the autoencoder output as the best, which is inconsistent with subjective human evaluation.

\subsection{Different layers activations}

\begin{figure}[h]
\centering
\includegraphics[width=7.5cm]{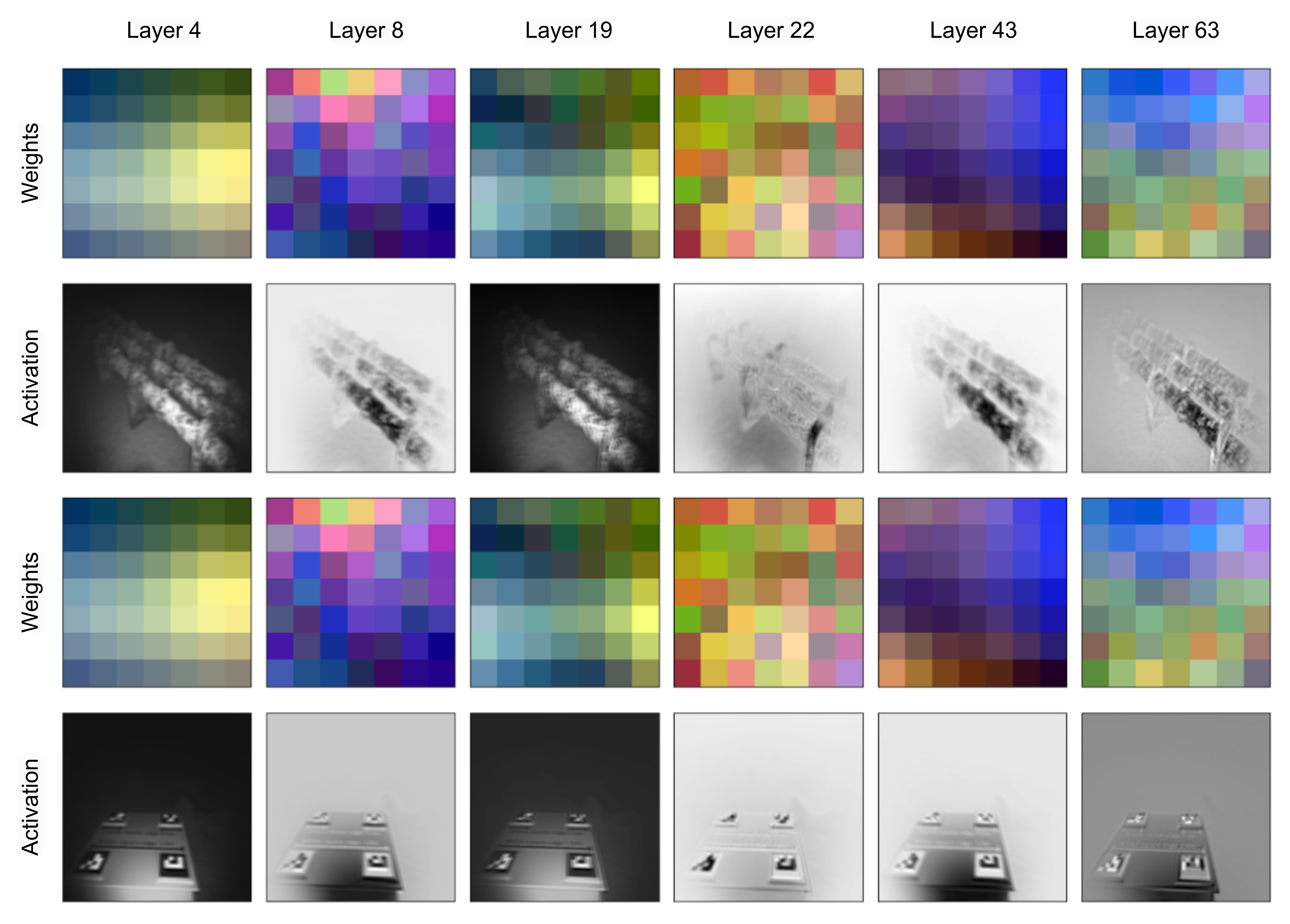}
  \caption{Visualization of activations and weights of the autoencoder model. We examine the weights and activated outputs of specific layers (4, 8, 19, 22, 43, and 63) in the autoencoder model. The filter sizes in these layers are 7 by 7.}
  \label{fig:filt}
\end{figure}

 We visualize the output from different layers of the autoencoder model and display these results in Figure \ref{fig:filt}. We hope that understanding each layer's learned features will give us a better idea of the model performance and training. To achieve this, we visualize the weights and the activated outputs from layers 4, 8, 19, 22, 43, and 63 of the autoencoder model. The filter sizes are all 7 by 7. The weight provides valuable insights into the learned features of each layer. For example, Layers 8 and 43 play a crucial role in extracting the light attenuation characteristics of underwater images.

 \subsection{Failure cases in unpaired image translation}

\begin{figure}[h]
\centering
\includegraphics[width=7.5cm]{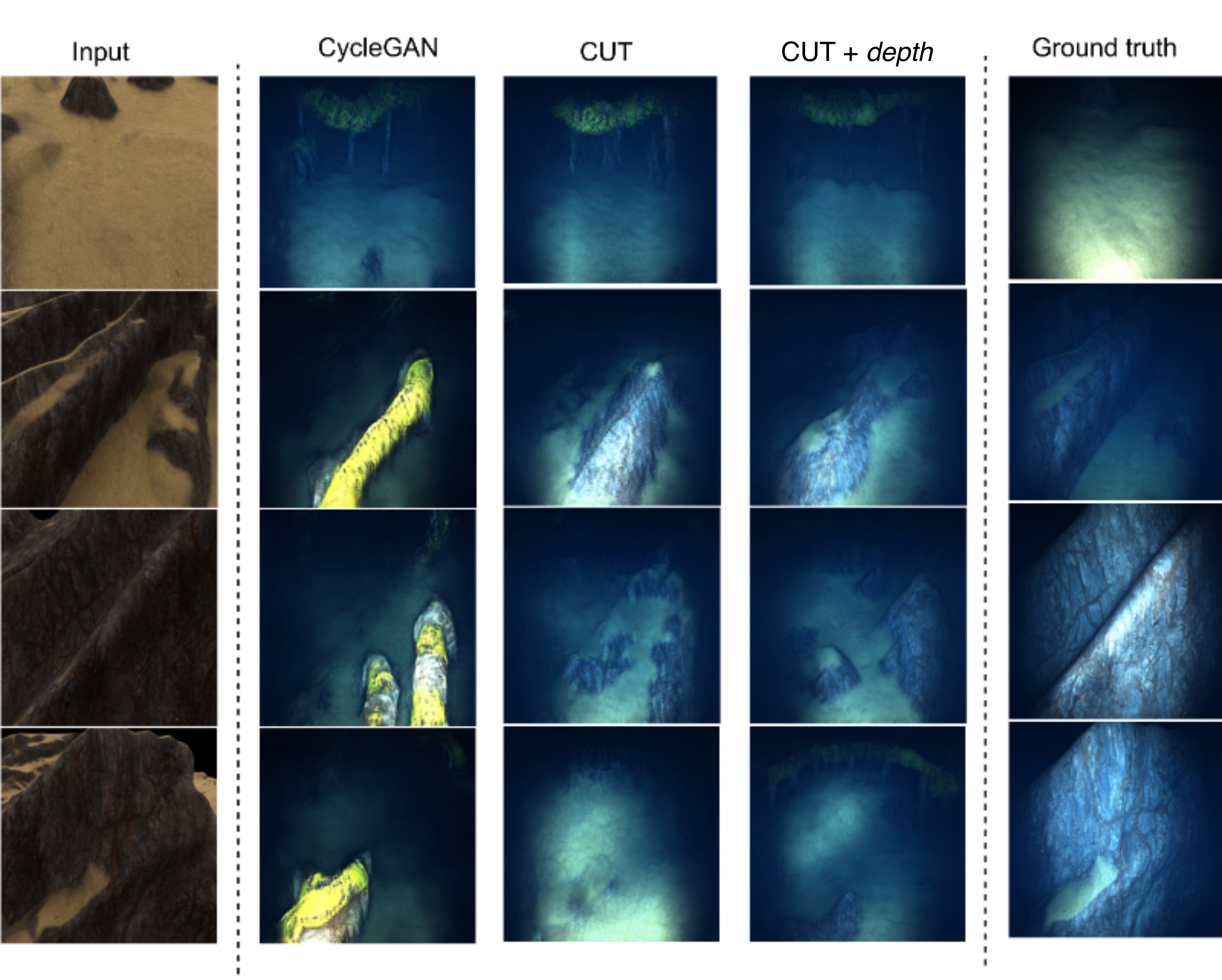}
  \caption{Typical failure cases of unpaired image translation techniques on underwater image inputs with scenes that contain confusing structures. CycleGAN \cite{zhu2017unpaired} generates images with color and texture of pipelines from a uniform lighting input without pipelines.}
  \label{fig:failcases}
\end{figure}

 A significant shortcoming of using GANs for generating realistic underwater images is that although the model learns to transfer the underwater style effectively, it struggles to preserve the content of the input image when the test examples come from a distribution different from the training data or contain structures that could be confused with another. This is apparent in the failure cases shown in Figure \ref{fig:failcases}.

 \subsection{Replacing the VAROS Synthetic Underwater Data}

\begin{figure}[h]  
  \centering
  \includegraphics[width=8.0cm]{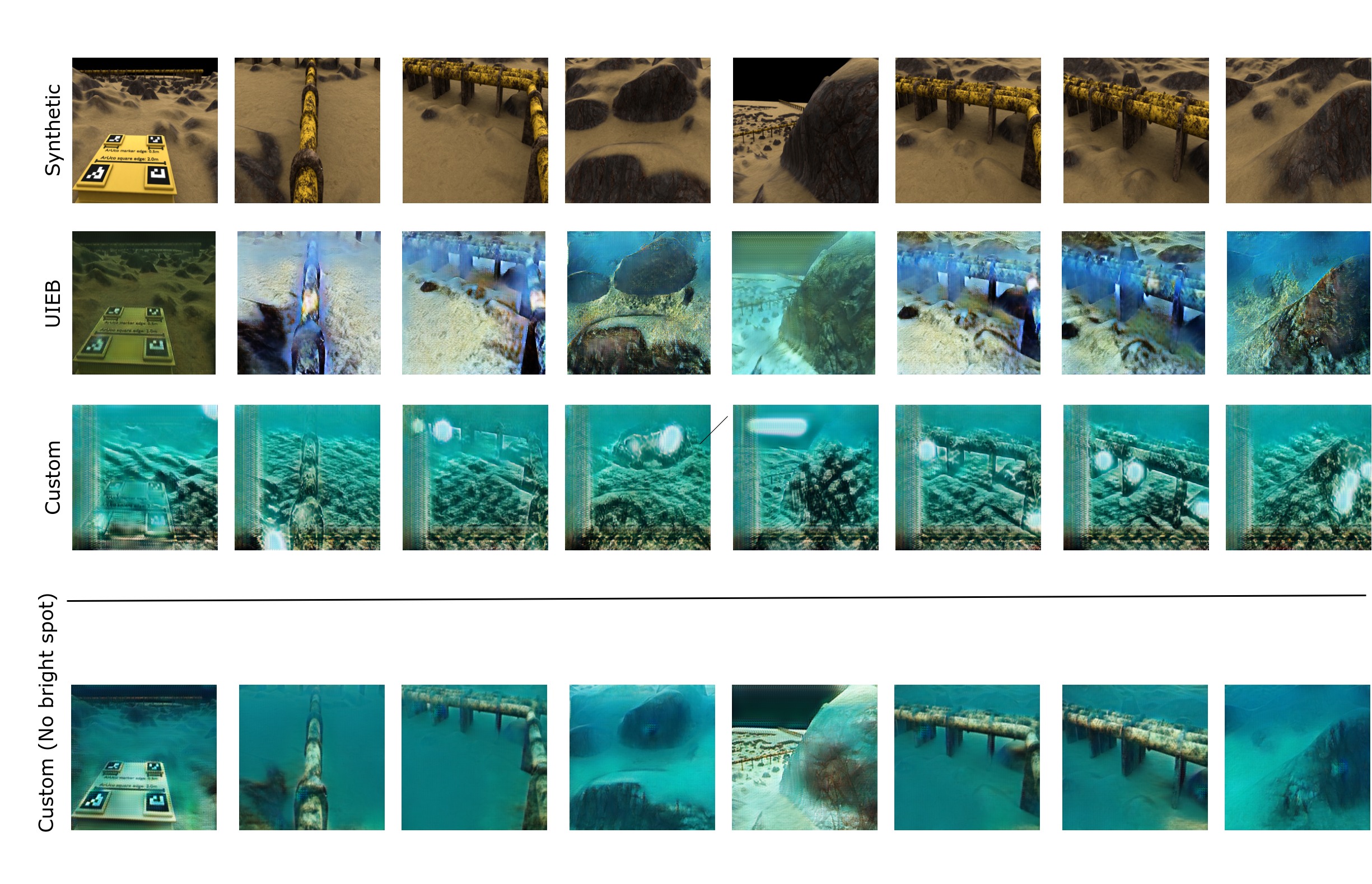}
  \caption[Additional real underwater dataset results]{The result obtained training the CUT model using the synthetic input from the VAROS underwater dataset conditioned on real images from the UIEB dataset.}
  \label{fig:customuieb}
\end{figure}

 We explored other variants of real underwater images to condition the unpaired image translation CUT model rather than just the VAROS underwater dataset. It is important to note that in all the cases, the synthetic uniform lighting input is still from the VAROS dataset. We observe that while the model effectively captures underwater properties, artifacts like coral reefs and sunlight introduce distortions that degrade the expected output. The results obtained upon conditioning on real underwater images from the UIEB \cite{uieb} dataset are shown in Figure \ref{fig:customuieb}. 

\section{Discussion}

The results in Table \ref{tab:eval} provide insights into the performance of different image translation models on the VAROS underwater dataset using two key evaluation metrics: Fréchet Inception Distance (FID) and Structural Similarity Index Measure (SSIM). For paired image translation, the pix2pix model achieves the best FID score (206) on the 6-sample subset and remains competitive (128) on the full 190-test set. This aligns with expectations, as pix2pix leverages paired supervision and the PatchGAN discriminator, which enhances fine-grained details. However, the autoencoder model achieves the highest SSIM (0.70) on the 6-sample test, suggesting that its reconstructions preserve more structural similarity to the input images despite a significantly higher FID (314). This discrepancy highlights a common limitation of autoencoders—while they optimize for pixel-wise similarity (MSE loss), they often produce blurry reconstructions, leading to higher FID scores. For unpaired methods, CycleGAN achieves the second-best FID (108) on the full test set, outperforming both CUT and the autoencoder. This result suggests that the cycle-consistency loss in CycleGAN provides effective supervision for mapping between domains. However, CUT, which replaces cycle-consistency loss with contrastive learning, achieves a slightly higher SSIM (0.69 vs. 0.67), indicating that it retains better overall structural similarity despite a slightly worse FID (132 vs. 108). The addition of depth information in CUT + depth results in the best overall FID score (106) on the full test set, demonstrating that depth cues improve the realism of generated underwater images. However, its SSIM score (0.66) is slightly lower than standard CUT (0.69), suggesting that while depth helps generate visually realistic images, it might introduce small structural distortions or variations that deviate from the reference images.

{
    \small
    \bibliographystyle{ieeenat_fullname}
    \bibliography{main}
}


\end{document}